\documentclass[letterpaper, 10 pt, journal, twoside]{IEEEtran}

\usepackage{graphics} 
\usepackage{epsfig} 
\usepackage{amsmath} 
\usepackage{amssymb}  

\usepackage[caption=false, font=footnotesize]{subfig}
\usepackage{algorithm}
\usepackage{algpseudocode}
\algrenewcommand\algorithmicrequire{\textbf{Input:}}
\algrenewcommand\algorithmicensure{\textbf{Output:}}
\newtheorem{assumption}{Assumption}
\newtheorem{definition}{Definition}
\usepackage{xcolor}
\usepackage[normalem]{ulem}

\usepackage[absolute,overlay]{textpos}
\usepackage{hyperref}

\title{
Sim-to-Real Surgical Robot Learning and Autonomous Planning for Internal Tissue Points Manipulation using Reinforcement Learning
}

\author{Yafei Ou and Mahdi Tavakoli, \textit{Senior Member, IEEE}
\thanks{Manuscript received: October 26, 2022; Revised: January 17, 2023; Accepted: February 28, 2023.}
\thanks{This paper was recommended for publication by Editor Jessica Burgner-Kahrs upon evaluation of the Associate Editor and Reviewers' comments. This work was supported by the Natural Sciences and Engineering Research Council (NSERC) of Canada and the China Scholarship Council (CSC). \textit{(Corresponding author: Yafei Ou)}}
\thanks{Yafei Ou and Mahdi Tavakoli are with the Department of Electrical and Computer Engineering, University of Alberta, Edmonton, Alberta, Canada. {\tt\footnotesize \{yafei.ou, mahdi.tavakoli\}@ualberta.ca}}%
\thanks{Digital Object Identifier (DOI): see top of this page.}
}

\begin{document}

\newcommand{\DOI}{10.1109/LRA.2023.3254860}

\thispagestyle{plain}
\onecolumn
\pagenumbering{gobble}

{
    \topskip0pt
    \vspace*{\fill}
    \LARGE
    \begin{center}
        © 2023 IEEE.
    \end{center}
    \vspace{1cm}
    \noindent Personal use of this material is permitted.  Permission from IEEE must be obtained for all other uses, in any current or future media, including reprinting/republishing this material for advertising or promotional purposes, creating new collective works, for resale or redistribution to servers or lists, or reuse of any copyrighted component of this work in other works. \\ 
    \vspace{1cm}
    \begin{center}
        DOI: \DOI
    \end{center}
    \vspace*{\fill}
}

\twocolumn 
\pagenumbering{arabic}
\begin{textblock*}{8cm}(0.541in,0.52in) 
   \href{https://doi.org/10.1109/LRA.2023.3254860}{\footnotesize \textcolor[HTML]{000080}{https://doi.org/10.1109/LRA.2023.3254860}}
\end{textblock*}

\maketitle

\begin{abstract}
Indirect simultaneous positioning (ISP), where internal tissue points are placed at desired locations indirectly through the manipulation of boundary points, is a type of subtask frequently performed in robotic surgeries. Although challenging due to complex tissue dynamics, automating the task can potentially reduce the workload of surgeons. This paper presents a sim-to-real framework for learning to automate the task without interacting with a real environment, and for planning preoperatively to find the grasping points that minimize local tissue deformation. A control policy is learned using deep reinforcement learning (DRL) in the FEM-based simulation environment and transferred to real-world situation. Grasping points are planned in the simulator by utilizing the trained policy using Bayesian optimization (BO). Inconsistent simulation performance is overcome by formulating the problem as a state augmented Markov decision process (MDP). Experimental results show that the learned policy places the internal tissue points accurately, and that the planned grasping points yield small tissue deformation among the trials. The proposed learning and planning scheme is able to automate internal tissue point manipulation in surgeries and has the potential to be generalized to complex surgical scenarios.
\end{abstract}
\begin{IEEEkeywords}
Medical robots and systems, reinforcement learning, dual arm manipulation, task and motion planning, surgical automation.
\end{IEEEkeywords}

\section{Introduction}
\IEEEPARstart{D}{ue} to its enhanced accuracy and dexterity, robot-assisted surgery (RAS) is becoming increasingly popular in surgical practice. While existing robotic surgery systems provide low-level assistance such as tremor reduction, recent research has focused on automating common surgical subtasks to achieve a higher level of autonomy \cite{chen2016virtual,chiu2021bimanual,richter2021autonomous}. One of the most frequently performed subtasks is deformable tissue manipulation, which may include placing tissue points at specific locations during needle insertion or retracting the tissue to reveal the region of interest underneath it. However, the automation of tissue manipulation presents a much greater challenge than many other subtasks since it involves a wide range of complex tissue dynamics that are difficult to model. Additionally, soft tissue manipulation in the surgical setting often demands high accuracy, which makes it more challenging than manipulating general soft objects.

\begin{figure}[t]
\centering
\includegraphics[width=3in]{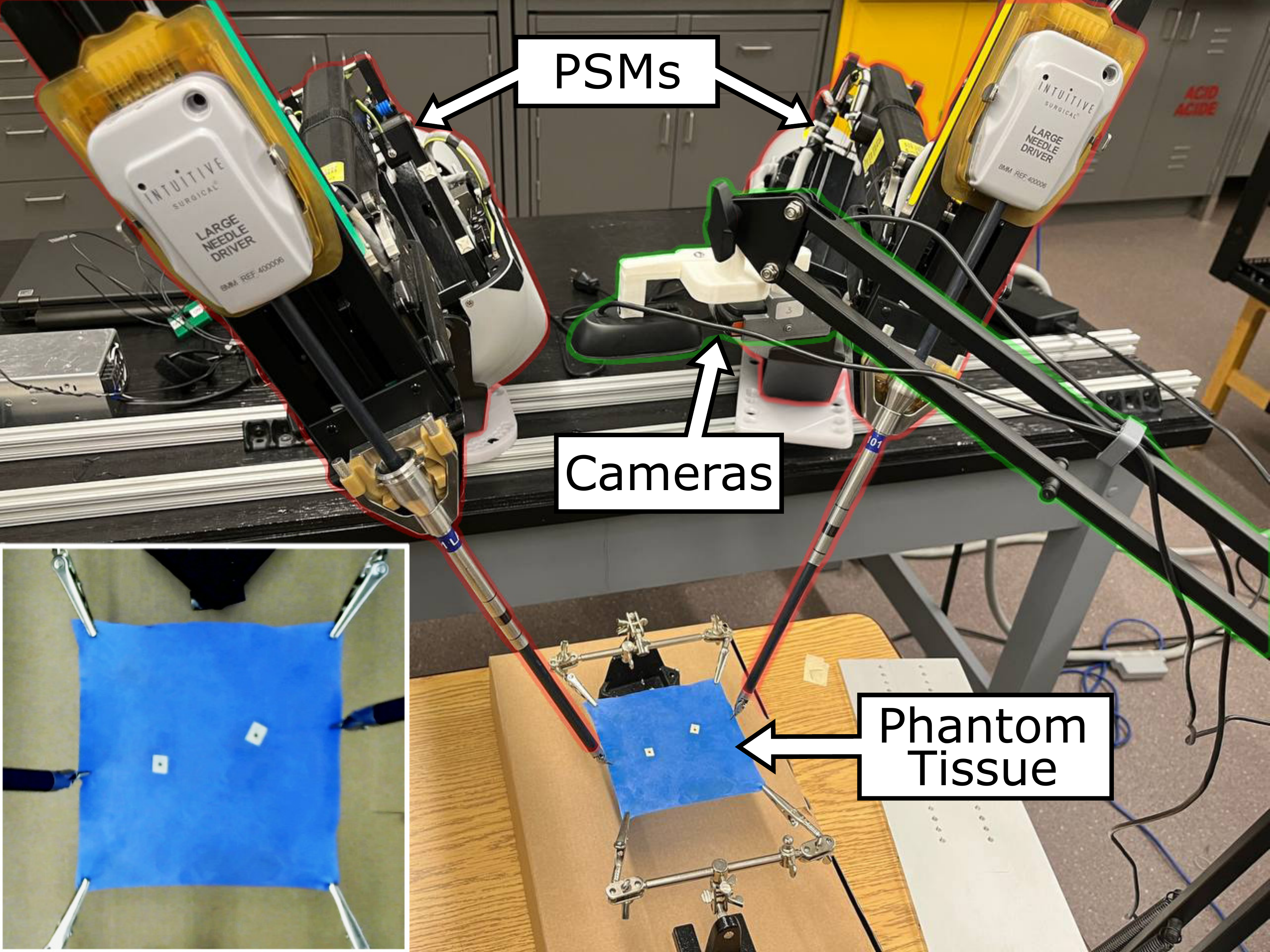}
\caption{Experimental setup for internal tissue point manipulation with two Patient Side Manipulators (PSMs, red) and a stereo camera system (green).} \label{fig:real_setup}
\end{figure}

There are two major areas that contribute to the advancement of automating surgical tissue manipulation: perception and task automation. Perception for surgical tissue manipulation aims at understanding the surgical situation based on vision inputs during the surgery, such as an endoscopic camera. This may include the detection, and tracking of surgical tools, as well as 3D reconstruction of the deformed tissue \cite{li2020super,lu2021super,lin2022semantic}. Task automation, on the other hand, involves automating the surgical robot to accomplish specific manipulation tasks, given the knowledge of the surgical situation. While perception serves as the basis of most task automation and control algorithms, the focus of this work is on the problem of high-level task automation.

Existing methods for controlling and automating surgical tissue manipulation can be classified into model-based and data-driven approaches. Model-based methods exploit human knowledge about the physical properties of the tissue to build an accurate model for predicting and controlling its deformation. Zhong {et al.} \cite{zhong2019dual} proposed an autonomous needle insertion scheme in which a robot manipulates the tissue to align a point with the target needle-tip position using finite element modelling (FEM) of the tissue. Similarly, we used model predictive control (MPC) by establishing an accurate FEM model for manipulating internal points of a breast tissue \cite{afshar2022model}. However, accurate tissue model parameters are usually patient specific and not always available prior to the surgery. Data-driven approaches, on the other hand, rely on trial and error to learn to control the tissue. For example, Alambeigi {et al.} \cite{alambeigi2018toward} used an online optimization scheme to learn the deformation Jacobian during bimanual kidney tissue manipulation.

Deep reinforcement learning (DRL) is a data-driven approach that has shown promising results in the field of robotic automation, including the manipulation of general deformable objects \cite{wu2019learning}. Shin {et al.} \cite{shin2019autonomous} proposed a model-based reinforcement learning approach for surgical tissue manipulation. However, this approach requires a large number of interactions with the environment (``explorations"), which is impractical in real surgical settings. To address this issue, recent works on autonomous surgery exploit a sim-to-real approach, which trains a policy in a simulation environment and transfers it directly to the real world. One option to ensure the learned policy is generalizable to the real world is to make the simulation environment as close as possible to reality. However, this again demands accurate physical model. Another solution to bridging the ``reality gap'' is domain randomization \cite{tobin2017domain}, where the simulation environment is randomized to cover a large range of properties during training, so that the learned policy works across a domain that includes the real world environment.

There have been promising results with sim-to-real DRL in the automation of some surgical tasks, such as cutting \cite{thananjeyan2017multilateral} and needle passing \cite{chiu2021bimanual}. Regarding tissue manipulation, a simulation environment has been developed based on position-based dynamics (PBD) in \cite{tagliabue2020soft}, where a policy for grasping and retracting the tissue to reveal a region of interest is learned and transferred to a real-world setup. In \cite{liu2021real}, a real-to-sim registration approach for simulating deformable tissue based on PBD was proposed. Although these works demonstrate the capability of PBD-based soft tissue simulation, it demands much parameter tuning before achieving realistic and accurate simulation behavior, which limits the use of PBD-based simulation in some settings that require accurate localization and control of the tissue points.

\begin{figure}
\centering
    \subfloat[\label{fig:isp}]{%
      \includegraphics[height=1.2in]{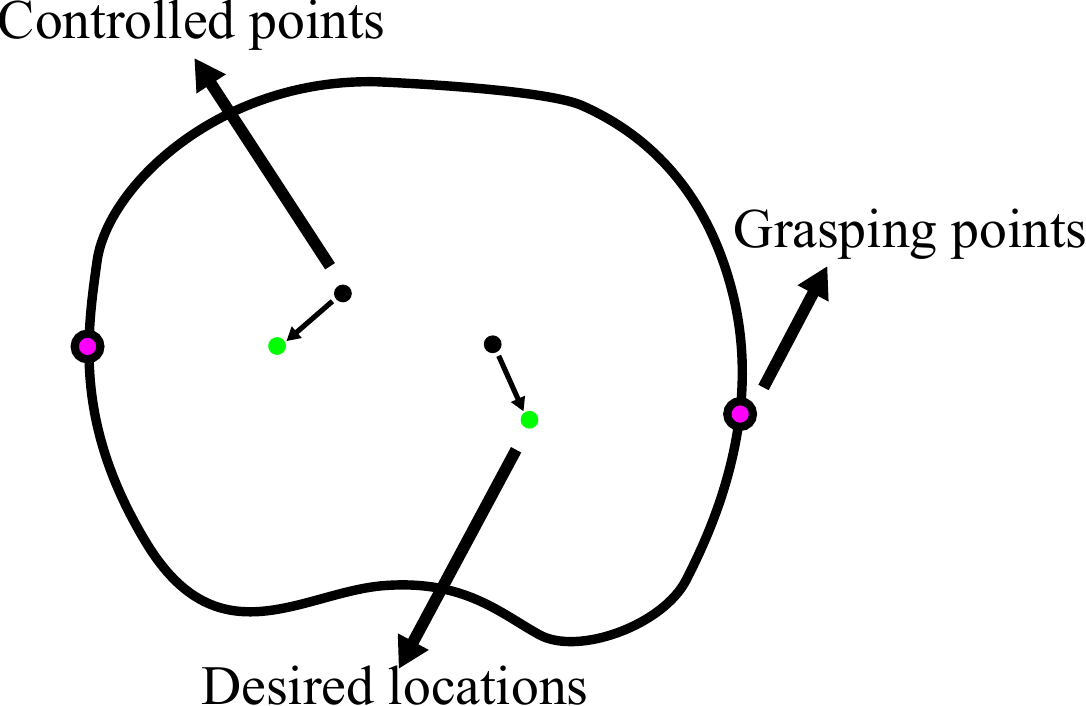}%
    }\hfil
    \subfloat[\label{fig:simulation}]{%
      \includegraphics[width=1.2in]{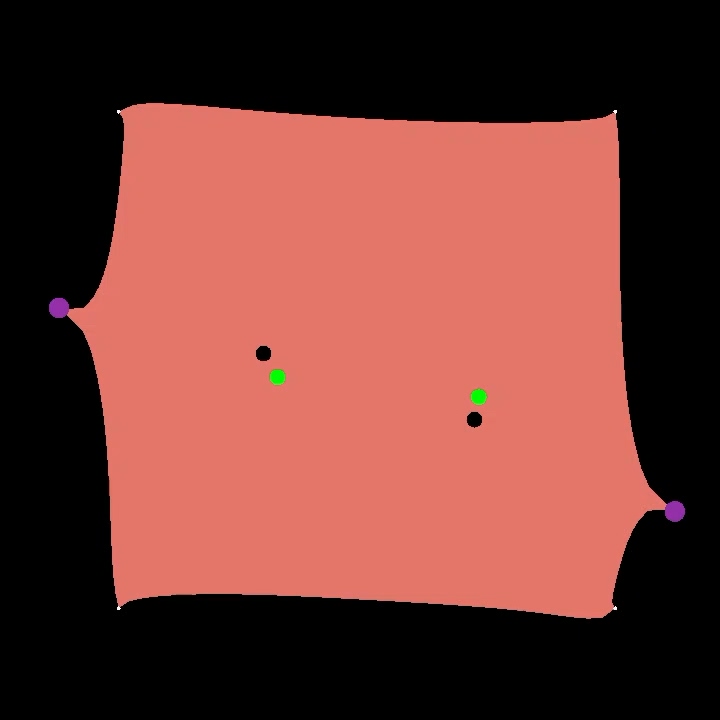}%
    }
    \caption{Indirect simultaneous positioning for tissue points: (a) problem definition; (b) FEM-based simulation environment. The controlled points and the desired positions are colored in black and green respectively, and the grasping points are marked in purple.}
    \label{fig:isp_simulation}
\end{figure}

\begin{figure*}[t]
\centering
\includegraphics[width=5in]{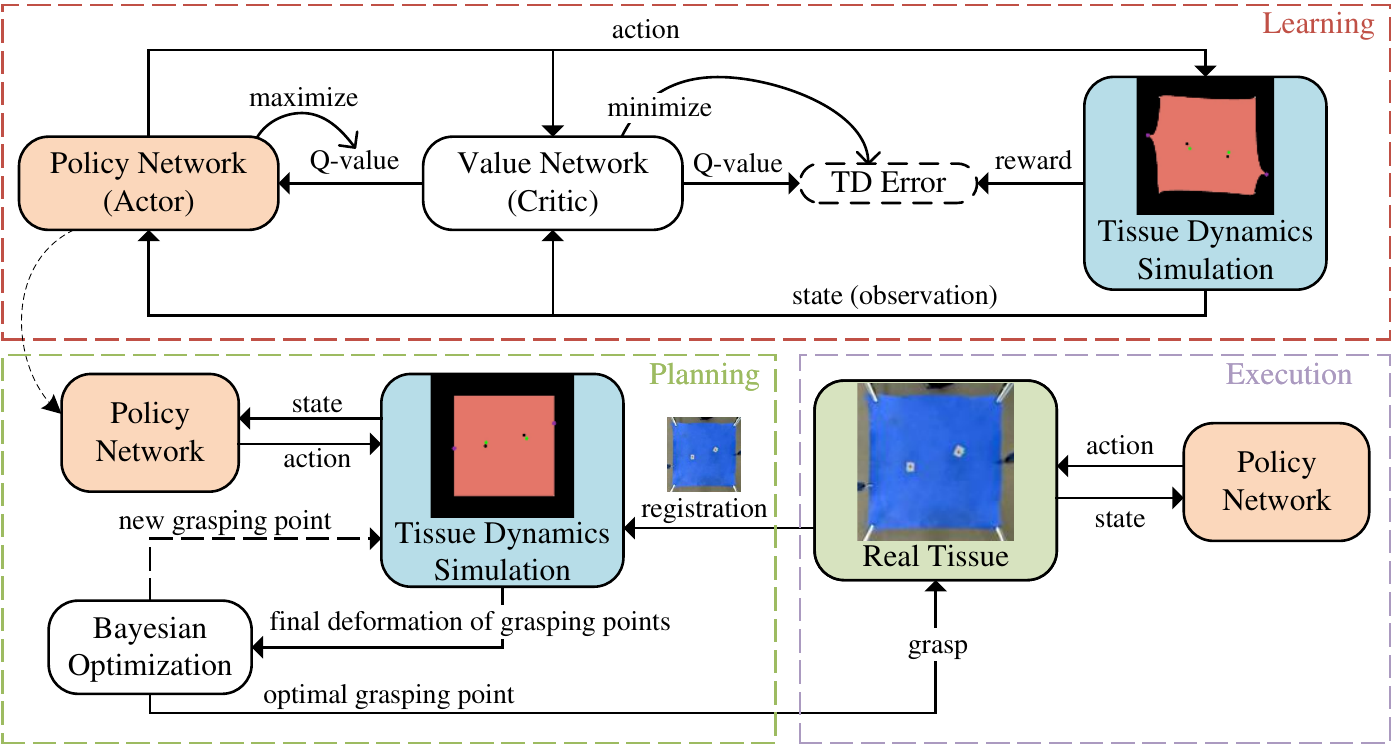}
\caption{Sim-to-real learning and planning for internal tissue points manipulation. {The policy is trained in the simulation environment during the learning phase, which is utilized to find the optimal grasping points during the planning phase. During the execution, the policy is able to automatically complete the task.} The state is the error between the current position of controlled points and the desired points $\mathbf{p}_{t} - \mathbf{p}_{des}$, and the action is the displacement of the grasping points $\Delta\mathbf{q} = \mathbf{q}_{t+1}-\mathbf{q}_{t}$.} \label{fig:workflow}
\end{figure*}

In this work, we focus on the indirect simultaneous positioning (ISP) problem \cite{wada1998indirect} for surgical tissue, where internal points (controlled points) of the tissue are placed at desired locations indirectly through the manipulation of boundary points (grasping points) of the tissue, as described in Fig.~\ref{fig:isp}. This subtask exists frequently in surgical scenarios such as needle insertion \cite{zhong2019dual} and kidney cryoablation \cite{alambeigi2018toward}, where specific tissue points must be placed at the desired locations through manipulation. This work presents a novel sim-to-real learning and planning scheme for automating the ISP problem in surgery. The main contributions can be summarized as follows:
\begin{itemize}
\item We build an FEM-based simulation environment for the ISP problem in surgery and train a transferable policy in the simulator using domain randomization.
\item Inconsistent simulation behavior caused by the explorations during training is addressed by formulating the problem as an augmented MDP.
\item Grasping points are planned preoperatively through Bayesian optimization (BO) in the simulation to minimize tissue deformation using the learned policy.
\item A sim-to-real learning and planning scheme for automating internal tissue points manipulation in surgery is implemented and validated on a real robotic setup.
\end{itemize}
The proposed framework is summarized in Fig.~\ref{fig:workflow}. To the best of our knowledge, this is the first time a transferable control policy for accurate tissue point localization has been learned directly from simulation and incorporated into simulation-based preoperative planning.

\section{Methods}
\subsection{FEM-Based Dynamic Simulation}
\label{sec:fem_simulation}
Since we are concerned with the accurate positioning of internal tissue points, high accuracy is essential for achieving sim-to-real learning. Furthermore, real-time simulation is important because the learning process involves numerous interactions with the simulation environment, and slow simulation leads to prolonged training time. Therefore, a trade-off between accuracy and computational cost must be made.

In general, Mass-spring system (MSS), FEM and PBD are the three main approaches used for real-time deformable object simulation. MSS is usually considered inaccurate when modelling large deformations, thus not suitable for this application. PBD-based simulation is fast and stable for simulating visually plausible object behaviors, but does not guarantee physical accuracy. Furthermore, the parameters of PBD are not physically meaningful and {require} manual adjustment before achieving a realistic simulation. Although slower than PBD, FEM-based simulation is typically more accurate, and the parameters for FEM are related to physical properties, making it straightforward to apply domain randomization in the FEM-based simulation (e.g. randomizing the Young's modulus). Therefore, we build our simulation environment on the SOFA simulator \cite{faure2012sofa}, an open-source FEM-based soft object simulation framework.

In FEM-based simulation, the system dynamics is modeled based on the Newton's law:
\begin{equation}
    \mathbf{M}\,\ddot{x}(\tau) = F_{int}(\tau) + F_{ext}(\tau)
\end{equation}
$\mathbf{M}$ is the mass matrix, $x$ is the degrees of freedom (DOFs) of the system, and $F_{int}$ and $F_{ext}$ are the internal and external forces applied to each DOF. To solve the system numerically, the Euler implicit (backward Euler) integration scheme is often used, approximating the system as
\begin{equation}
\begin{gathered}
    x(\tau+\Delta{\tau}) = x(\tau) + \Delta{\tau} \cdot v(\tau+\Delta{\tau}) \\
    \mathbf{M} (v(\tau+\Delta{\tau})-v(\tau))=\Delta{\tau}\cdot(F_{int}(\tau+\Delta{\tau}) + F_{ext}(\tau))
\end{gathered}
\end{equation}
where $v$ is the velocity and $\Delta{\tau}$ is the integration time step. We use the iterative conjugate gradient (CG) optimizer to solve the system iteratively. Unlike static FEM, dynamic FEM simulation continues to the next time step even if the iterative solver has not converged after a specified maximum number of iterations.

In this work, the tissue is modeled by a triangle mesh. To simulate the behavior of grasping and moving specific points of the tissue, we directly apply displacement constraints to the nodes that are considered as being grasped and moved. Spring forces are applied to the tissue nodes based on the distance between the target and the current position of each node:
\begin{equation}
    F_a = k_s(\mathbf{q}_{t+1}-\mathbf{q}_{t})
\end{equation}
where $k_s$ is a large stiffness coefficient manually defined, and $\mathbf{q}_{t}$ and $\mathbf{q}_{t+1}$ are the node positions before and after the control step at time $t$. Prior work shows that this approach has a good overall performance compared with two other methods that use contact detection \cite{tagliabue2020biomechanical}. Fig.~\ref{fig:simulation} shows the simulation environment.

\subsection{RL Problem Formulation}
A reinforcement learning (RL) problem is usually formulated as a Markov Decision Process (MDP) described by a tuple $(\mathcal{S}, \mathcal{A}, P, R, \gamma)$, where $\mathcal{S}$ and $\mathcal{A}$ are the state and action spaces, $P:\mathcal{S}\times \mathcal{A} \times\mathcal{S} \rightarrow [0,1]$ is the state transition function, $R:\mathcal{S}\times\mathcal{A}\rightarrow\mathbb{R}$ is the reward function, and $\gamma \in [0,1]$ is the discount factor. An agent learns to optimize a policy $\pi:\mathcal{S}\times \mathcal{A} \rightarrow [0,1]$ through the interaction with the environment to maximize the expected return 
\begin{equation}
\label{eqn:rl_expected_return}
    \pi^* = \mathop{\arg \max}_{\pi} \sum_{t=0}^{T} \mathbb{E}_{(\mathbf{s}_t, \mathbf{a}_t)\sim \rho_\pi}\left[ \gamma^{t} r(\mathbf{s}_t, \mathbf{a}_t) \right]
\end{equation}
where $\rho_\pi(\mathbf{s}_t, \mathbf{a}_t)$ is the distribution of the trajectory $(\mathbf{s}_1, \mathbf{a}_1, \dots, \mathbf{s}_T, \mathbf{a}_T)$ produced by the policy $\pi$.

In this work, we first formulate the problem as a regular MDP  $(\mathcal{S}, \mathcal{A}, P, R, \gamma)$, while we will discuss later that this simple approach does not yield good results. Although the actual state space is the position of all the tissue nodes, we approximate it with the difference between the position of the controlled points and the desired locations. The actions are the displacements of each grasping point from the current location in cartesian coordinates at each action step. The state and action spaces can be expressed mathematically as follows:
\begin{equation}
    \begin{gathered}
    \mathbf{s}_{t} = \mathbf{p}_{t} - \mathbf{p}_{des} \in \mathbb{R}^{N \cdot D}\\
    \mathbf{a}_{t} = \Delta\mathbf{q} = \mathbf{q}_{t+1}-\mathbf{q}_{t} \in [-0.2,0.2]^{M\cdot D}
    \end{gathered}
\end{equation}
where $N$ is the number of the controlled points, $M$ is the number of grasping points, $D$ is the dimension of the operating space, $\mathbf{p}_{t}$ represents the positions of the controlled points, $\mathbf{p}_{des}$ represents the desired positions, and $\Delta\mathbf{q}$ is the displacement of each grasping point from the current location, with a maximum of $0.2$ mm along each direction. In our case, $N=M=D=2$ since we consider a 2D space manipulation task of controlling two internal tissue points through the movement of two grasping points. The reward function is designed to reflect how close the controlled points are to the desired locations:
\begin{equation}
    r(\mathbf{s}_{t}, \mathbf{a}_{t}) = \lambda \left(1-\sqrt{\frac{\lVert \mathbf{p}_{t}-\mathbf{p}_{des}\rVert}{\lVert \mathbf{p}_{0}-\mathbf{p}_{des}\rVert}}\right)
\end{equation}
where $\mathbf{p}_{0}$ is the initial position of the controlled points and $\lambda$ is a scaling factor.

\subsection{Soft Actor-Critic}
Soft actor-critic (SAC) \cite{haarnoja2018soft} is an off-policy actor-critic algorithm that has recently shown promising results in learning continuous control problems and is known to be robust to environment changes. As will be discussed in Section~\ref{sec:inconsistent_performance}, changing environment is considered in this work, making SAC suitable for this application. It considers the maximum entropy reinforcement learning problem, which augments the standard reinforcement learning objective (\ref{eqn:rl_expected_return}) with an entropy term:
\begin{equation}
\label{eqn:rl_sac_obj}
    \pi^* = \mathop{\arg \max}_{\pi}\sum_{t=0}^{T} \mathbb{E}_{(\mathbf{s}_t, \mathbf{a}_t)\sim \rho_\pi}\left[ \gamma^{t} r(\mathbf{s}_t, \mathbf{a}_t) + \alpha \mathcal{H}(\pi(\cdot | \mathbf{s}_t)) \right]
\end{equation}
where $\mathcal{H}(\pi(\cdot | \mathbf{a}_t))$ is the entropy of the action distribution under the state $\mathbf{s}_t$, and $\alpha$ is temperature parameter determining the importance of the entropy term. This objective aims to maximize the expected return and the entropy of the action at the same time, which can encourage exploration and capturing multiple near-optimal actions \cite{haarnoja2018soft}.

SAC algorithm exploits an actor-critic policy search scheme, including a policy $\pi_\phi$ as the actor and a soft Q-function $Q_\theta$ as the critic. During learning, $Q_\theta$ is updated iteratively to approximate the temporal difference (TD) target:
\begin{equation}
    \hat{y}_t = r(\mathbf{s}_t, \mathbf{a}_t) + \gamma \mathbb{E}_{\mathbf{s}_{t+1}\sim p}[V_{\theta}(\mathbf{s}_{t+1}) ]
\end{equation}
where $V_{\theta}(\mathbf{s}_{t}) = \mathbb{E}_{\mathbf{a}_{t}\sim \pi}[Q_{\theta}(\mathbf{s}_{t}, \mathbf{a}_{t}) - \alpha \log(\pi(\mathbf{a}_t | \mathbf{s}_t))]$ is the state value function. Hence, $Q_\theta$ can be updated by minimizing the loss
\begin{equation}
    J_Q(\theta) = \mathbb{E}_{(\mathbf{s}_t, \mathbf{a}_t)\sim \mathcal{D}}\left[\frac{1}{2}\left( Q_\theta(\mathbf{s}_t, \mathbf{a}_t) - \hat{y}_t \right)^2 \right]
\end{equation}
and the policy $\pi_\phi$ can be updated by minimizing the loss
\begin{equation}
    J_\pi(\phi) = \mathbb{E}_{\mathbf{s}_t\sim \mathcal{D}}\left[ \mathbb{E}_{\mathbf{a}_t\sim \pi_\phi} \left[ -Q_\theta(\mathbf{s}_t, \mathbf{a}_t) + \alpha \log(\pi(\mathbf{a}_t | \mathbf{s}_t)) \right]  \right]
\end{equation}
where $\mathcal{D}$ is the experience replay buffer.

\subsection{Learning with Inconsistent FEM Performance}
\label{sec:inconsistent_performance}
As discussed in Section \ref{sec:fem_simulation}, the dynamic simulation continues to the next time step even if the iterative solver has not converged if the maximum allowed iteration number is reached. While relatively good simulation performance is guaranteed by manually choosing a suitable maximum iteration number through experiments, it degrades occasionally during DRL training and induces significant calculation delay in solving FEM. As a result, the response of controlled points falls largely behind the movement of the grasping points in this case. This is related to the two factors discussed in the following.

First, the agent is encouraged to explore the environment sufficiently by including randomness when taking actions. Although this is a standard procedure, the repetitive movements of the grasping points results in large internal force $F_{int}$ remaining in the simulated object between each simulation step that continues to take effect at the next simulation step, leading to the aforementioned calculation delay.

Second, while an expert policy completes the task by taking only a few actions, the randomly initialized DRL policy performs badly and takes a large number of actions during one single episode. The error of the internal force $F_{int}$ is likely to accumulate, especially when large movements of the grasping points are taken consecutively.

To formally define the problem, the following assumption is made on the dynamic FEM simulation environment based on intuition:
\begin{assumption}The position of the controlled points $\mathbf{p}_{ctr}$ is dependent on the $n$ most recent actions, where $n$ is a variable no larger than a constant $K$.
\end{assumption}

Since the actual value of $K$ is not known, $K$ is considered to be a hyper-parameter manually chosen in this work. Based on Assumption 1, we have a time-varying transition function
\begin{equation}
\begin{aligned}
    P_t(\mathbf{s}_{t+1}|\mathbf{s}_t,\mathbf{a}_t,\mathbf{s}_{t-1},\mathbf{a}_{t-1},\dots, \mathbf{s}_0,\mathbf{a}_0) \\ = P_t(\mathbf{s}_{t+1}|\mathbf{s}_t,\mathbf{a}_t, \mathbf{a}_{t-1}, \dots, \mathbf{a}_{t-n+1})
\end{aligned}
\end{equation}
Therefore, the state transition depends on not only the current action but also previous ones when $n > 1$, violating the Markov property. We have
\begin{equation}
    P_t(\mathbf{s}_{t+1}|\mathbf{s}_t,\mathbf{a}_t,\mathbf{s}_{t-1},\mathbf{a}_{t-1},\dots, \mathbf{s}_0,\mathbf{a}_0) =P_t(\mathbf{s}_{t+1}|\mathbf{s}_t,\mathbf{a}_t)
\end{equation}
which satisfies the Markov property only when $n=1$.

In other words, the Markov property is not always satisfied due to the simulation performance degradation caused by exploration, making it difficult for the original DRL algorithm to learn an optimal policy. While increasing the maximum number of iterations or including viscosity in the tissue model can potentially solve or mitigate this issue, a significant amount of time will be spent on solving the FEM simulation, leading to prolonged training time.

To address the issue, we choose the maximum iteration number to be as small as possible while ensuring that the simulation performance is good enough when a hand-crafted expert policy is manipulating the tissue in the simulation environment{, where the movement of each grasping point is related to the direction pointing from the closest controlled point to its desired location}. Furthermore, the following assumption is made based on the previous discussions.

\begin{assumption} A better policy with less randomness results in faster convergence of the FEM calculation. If the policy is close to optimal (e.g. an expert policy), the dynamic FEM calculation converges within one step, i.e. $n=1$.
\end{assumption}

Similar to \cite{altman1992closed} which formulates an augmented MDP problem to solve RL problems with observation delay, we can now construct a new state augmented MDP problem which always satisfies the Markov property. Let $(\mathcal{J}, \mathcal{A}, \{T_t\}, R, \gamma)$ be its state space, action space, transition function, reward function and discount factor, where
\begin{equation}
\begin{gathered}
\mathcal{J} := \mathcal{S}\times\mathcal{A}^{K-1} \\
\mathbf{j}_t := (s_t, a_{t-1}, a_{t-2}, \dots, a_{t-K+1})
\end{gathered}
\end{equation}
$\mathcal{J}$ is the augmented state space. Then the transition function is
\begin{equation}
\begin{aligned}
    T_t(\mathbf{j}_{t+1}|\mathbf{j}_t,\mathbf{a}_t) = P_t(\mathbf{s}_{t+1}|\mathbf{s}_t,\mathbf{a}_t, \mathbf{a}_{t-1}, \dots, \mathbf{a}_{t-K+1})
\end{aligned}
\end{equation}
The transition function $T_t$ is also time-varying, but the Markov property is satisfied.

So far, we have modeled the problem as an MDP with changing a transition function. In \cite{csaji2008value}, the authors propose modelling the problem of RL with changing transition function or cost function by an $(\epsilon,\delta)$-MDP:
\begin{definition}
A tuple $(\mathcal{S}, \mathcal{A}, \{P_t\}, \{R_t\}, \gamma)$ is called an $(\epsilon,\delta)$-MDP with $\epsilon,\delta > 0$ if there exists a \textit{base MDP} such that the change of the transition $\{P_t\}$ and cost function $\{R_t\}$ are asymptotically bounded by $\epsilon$ and $\delta$:
\begin{equation}
    \begin{aligned}
        \limsup_{t\rightarrow \infty}{\lVert P-P_t \rVert \le \epsilon} \\
        \limsup_{t\rightarrow \infty}{\lVert R-R_t \rVert \le \delta}
    \end{aligned}
\end{equation}
\end{definition}

\begin{assumption}
There exists a \textit{base MDP} $(\mathcal{J}, \mathcal{A}, T, R, \gamma)$ for the augmented MDP $(\mathcal{J}, \mathcal{A}, \{T_t\}, R, \gamma)$, such that
\begin{equation}
    \limsup_{t\rightarrow \infty}{\lVert T-T_t \rVert \le \epsilon}
\end{equation}
with $\epsilon > 0$. Thus, It can be modeled by an $(\epsilon,\delta)$-MDP.
\end{assumption}

According to \cite{csaji2008value}, directly applying standard value iteration algorithms to an $(\epsilon,\delta)$-MDP results in convergence to the optimal solution of the base MDP. Therefore, if Assumptions 1 to 3 are satisfied, applying the SAC algorithm to the MDP with state augmentation should also result in learning an optimal policy by the end. We demonstrate through experiment in the next section that these assumptions are reasonable and this training scheme can learn transferable policies efficiently from imperfect FEM-based simulations.

\subsection{Preoperative Planning of Grasping Points}
While the deformation of the tissue is not taken into account during training in the simulation, large deformation can cause damage to the soft tissue in real surgical practice and should be avoided.
To minimize the final tissue deformation after the internal points have been successfully placed at the desired locations, we further automate the task by finding the best grasping points based on the learned policy. Considering the local tissue deformation around the grasping points, an optimization problem is formulated as minimizing final displacement of the grasping points relative to their initial positions:
\begin{equation}
\begin{aligned}
    \mathbf{q}^{*}_{0} =\mathop{\arg \min}_{\mathbf{q}_{0} \in U}  f(\mathbf{q}_{0}) =\mathop{\arg \min}_{\mathbf{q}_{0} \in U} \left[\mathbf{q}_{e}(\mathbf{q}_{0}, \mathbf{p}_{0}, \mathbf{p}_{des}, \pi) -\mathbf{q}_{0}\right]
\end{aligned}
\end{equation}
where $\pi$ is the learned policy, $\mathbf{q}_{0}$ and $\mathbf{p}_{0}$ are the initial position of the grasping points and the controlled points, $\mathbf{p}_{des}$ is the desired position, and $U$ is a set containing all possible initial grasping locations. The final position of the grasping points $\mathbf{q}_{e}$ is therefore an implicit function of $\mathbf{q}_{0}$, $\mathbf{p}_{0}$, $\mathbf{p}_{des}$ and $\pi$. Since the evaluation of the objective function can only be achieved by acting in the simulation environment using the learned policy, Bayesian optimization (BO), which is known to be efficient in optimizing black-box objective functions, is chosen for solving the optimization problem.

The general procedure of preoperative grasping point optimization is summarized in Algorithm~\ref{alg:grasping_planning}.

\newcommand{\WRP}{\par\qquad\(\hookrightarrow\)\enspace}
\begin{algorithm}[hbt!]
\caption{Preoperative planning of grasping points}
\label{alg:grasping_planning}
\begin{algorithmic}
\Require{$\mathbf{p}_{0}$, $\mathbf{p}_{des}$, RL policy $\pi$, Gaussian process estimator (GP), acquisition function $a(\mathbf{q}_{0})$, hyper-parameters $n_0$, $N$}
\Ensure{$\mathbf{q}_{0}^{*}$}
\State Register $\mathbf{p}_{0}$, $\mathbf{p}_{des}$ in the simulator\;
\State Sample $n_{0}$ grasping point configurations $\mathbf{q}_{0}^{1:n_{0}}$\;
\State Evaluate $f$ with each $\mathbf{q}_{0}$ using policy $\pi$ in the simulation\;
\State $n \gets n_0$\;
\While{$n \le N$}
    \State Update the posterior distribution of GP \;
	\State Optimize $\mathbf{q}_{0}^{n} = \arg\max a(\mathbf{q}_{0})$\;
	\State Evaluate $f$ at $\mathbf{q}_{0}^{n}$ using the policy $\pi$ in the simulation\;
	\State $n \gets n+1$\;
\EndWhile
\State Choose $\mathbf{q}_{0}^{*}$ to be the the point with the smallest $f(\mathbf{q}_{0})$
\end{algorithmic}
\end{algorithm}

\begin{figure}[t]
\centering
\includegraphics[width=2.5in]{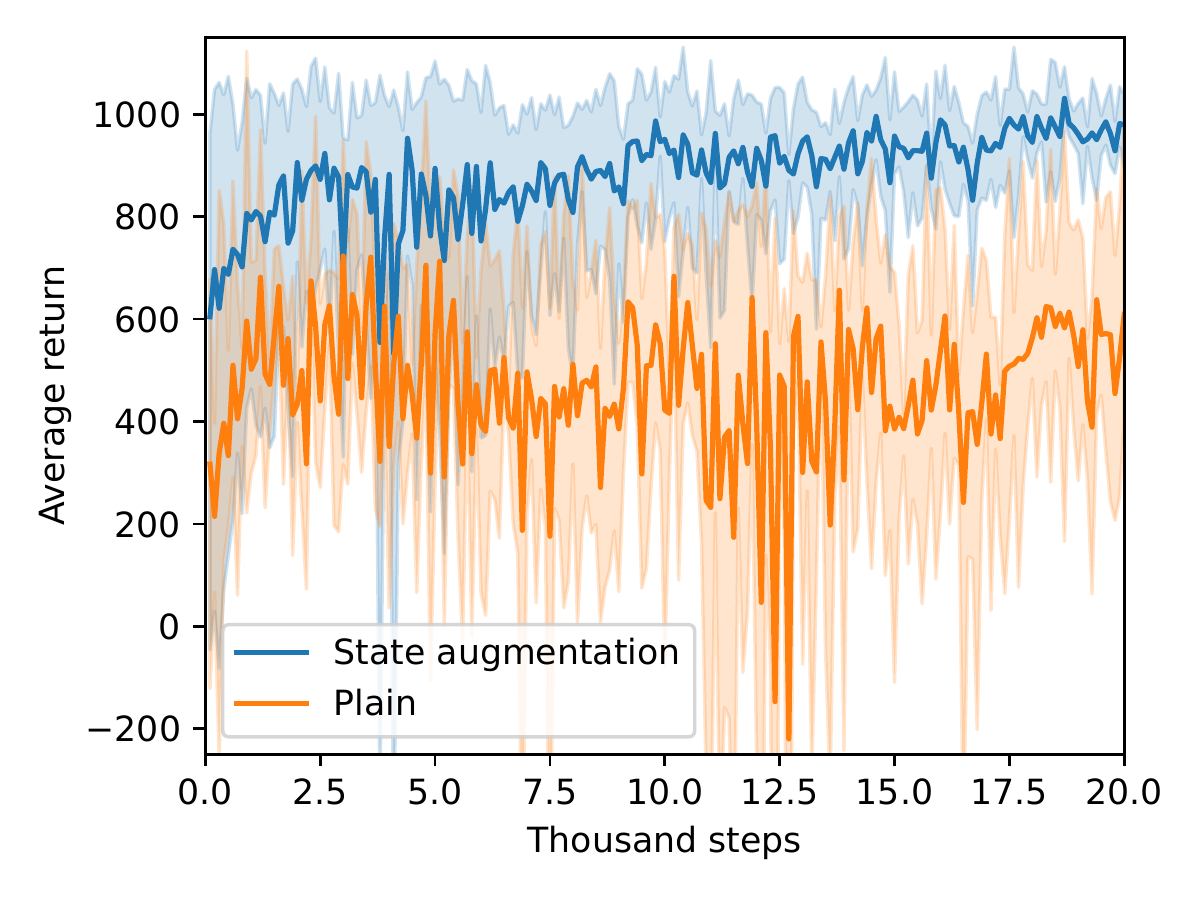}
\caption{Learning curve of the RL agents. The solid lines are the mean values. The shaded areas represent the standard deviation.} \label{fig:learning_curve}
\end{figure}

\section{Experiments and Results}
\subsection{Experiments}
\subsubsection{Training in the Simulation} Our DRL framework is implemented based on the stable-baselines 3 library \cite{stable-baselines3}. The DRL agent is trained in the simulation in an episodic manner. Each episode includes 100 action steps, and the parameters of the environment are reinitialized at the beginning of each episode. At each episode, the Young’s modulus of the simulated tissue is randomly sampled from $[0.6 \text{ MPa}, 1.2 \text{ MPa}]$, while the Poisson's ratio is kept to be $0.49$. The stiffness coefficient $k_s$ is $10^4$, and the maximum allowed number of iterations is 50. The grasping points are uniformly chosen from two predefined sets of valid grasping points. The controlled points are randomly selected from the central area of the tissue, and their corresponding desired locations are generated randomly around the initial locations of the controlled points with a distance of $ 4 \text{ mm}$. A resampling scheme is implemented to prevent the generated desired points from being too close to each other. This domain randomization technique ensures that the trained policy has good generalizability and is transferable to the real world. A PC with Intel(R) Core(TM) i5-9600K CPU is used for both the FEM simulation and the RL training process. GPU and parallelism are not employed in this work.

In this work, the actor and the critic networks are two-layer multilayer perceptron (MLP) networks with 256 hidden units at each layer. Since we assume that the simulation environment is changing gradually, a small experience replay buffer with its size being 5000 is used. The replay buffer is filled prior to the start of training by exploring with a randomly initialized policy. The agent is trained for 20 thousand steps with a batch size of 256 and a learning rate of 7e-4 for both the actor and the critic networks.

$K=5$ is selected for state augmentation. Two agents are trained with and without state augmentation as a comparison. For each agent, we train three instances with different random seeds. During training, the policy is evaluated for 5 episodes every 100 steps and the average return is collected.

\subsubsection{Planning of the grasping points} In this work, the implementation of Bayesian optimization is based on the scikit-optimize library. Matern kernel is used for the Gaussian process estimator. The expected improvement (EI) is used as the acquisition function $a$. $n$ and $N$ are chosen as 5 and 20, respectively. During planning, early termination of the simulation is triggered when the step reward is greater than 10, indicating that the controlled points are close enough to the desired locations.

\subsubsection{Experimental setup} The real experimental setup is shown in Fig.~\ref{fig:real_setup}. The four vertices of the tissue are fixed by clamps. Two Patient Side Manipulators (PSMs) from the da Vinci Research Kit (dVRK) \cite{kazanzides-chen-etal-icra-2014} are used to grasp the tissue on two sides. During the manipulation, the motions of the end-effectors (EEs) are controlled in a 2D plane in the Cartesian space and the orientations are kept constant.

The manipulation object is a highly elastic dental dam (size $10\times 10 \text{ cm}^2$) made of latex that simulates a phantom tissue. According to \cite{santos2021new}, the Young's modulus of rubber dams ranges from $0.6 \text{ MPa}$ to $1.2 \text{ MPa}$, making it an appropriate material for simulating a number of human soft tissue such as chest skin, colon, and uterus \cite{singh2021mechanical}. Two stickers with black dots are attached to the surface of the dental dam as markers for the controlled points.

A stereo camera system with Logitech C270 and C525 webcams (Logitech International S.A., Lausanne, Switzerland), calibrated using Matlab Stereo Camera Calibrator, is used to register between the real world and simulation before the experiments by manually selecting the four vertices of the phantom tissue and the two controlled points in the images.
Each camera operates at a frame rate of 30 Hz and has a resolution of 640 x 480. 
After scene registration, the position of the controlled points is registered in the simulation environment for the planning of grasping points.
During the manipulation, the Cartesian positions of the points are estimated based on the pixel locations from one single camera. As the cameras are placed perpendicular to the tissue, this simplification is reasonable and avoids localization errors caused by 3D reconstruction if both cameras are used. Through stereo calibration and registration, it can be approximated that $1 \text{ px} \approx 0.43 \text{ mm}$ based on the actual size of the phantom tissue. Since the conversion from pixel to millimeter incurs errors, raw data in the pixel domain is reported in this work.

To evaluate the trained policy and test the proposed preoperative planning scheme in the real-world environment, we randomly choose 5 different configurations of the controlled points and desired locations. The change of the controlled points is made by simply re-attaching the markers on the phantom tissue. Different desired locations are generated randomly around the initial locations of the controlled points with a distance of 4 mm for each of the 5 configurations. For each configuration, the optimal grasping locations are found and augmented in the image as visual guidance. The human operator manually {grasps} the tissue at the planned grasping points through teleoperation before the autonomous manipulation starts.

Each configuration is subjected to ten manipulation trials, of which only one trial uses the optimal grasping locations. They are randomly chosen in the other 9 trials. During each trial, the robot manipulates the tissue for 30 steps and the pixel error at each control step is collected. The displacement of the grasping point is recorded at the end.

\begin{figure}[t]
\centering
\includegraphics[width=2.8in]{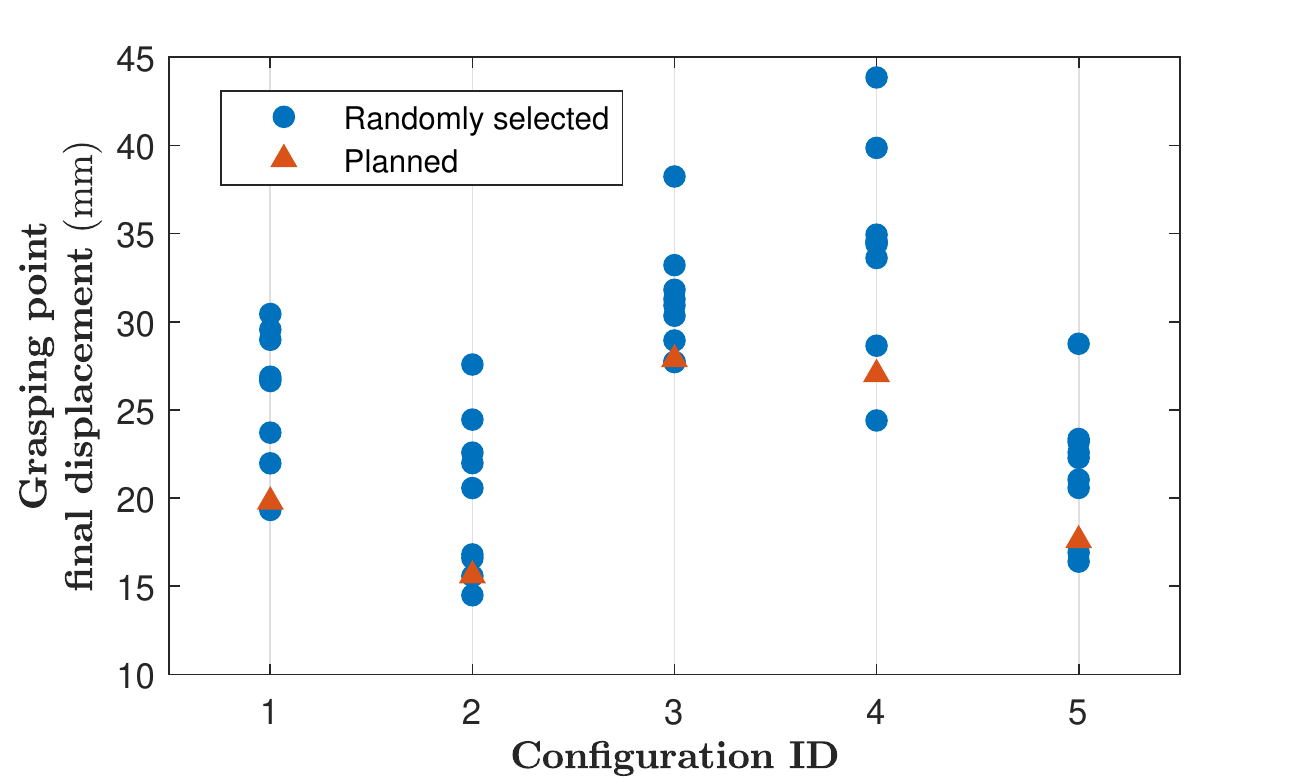}
\caption{Local tissue deformation at the grasping points after manipulation. The round dots represent the final displacement of the grasping points using randomly selected grasping points, and the triangular dots represent the final displacement using the planned grasping points.}
\label{fig:grasping_displacement}
\end{figure}

\begin{figure}[t]
\centering
\includegraphics[width=2.5in]{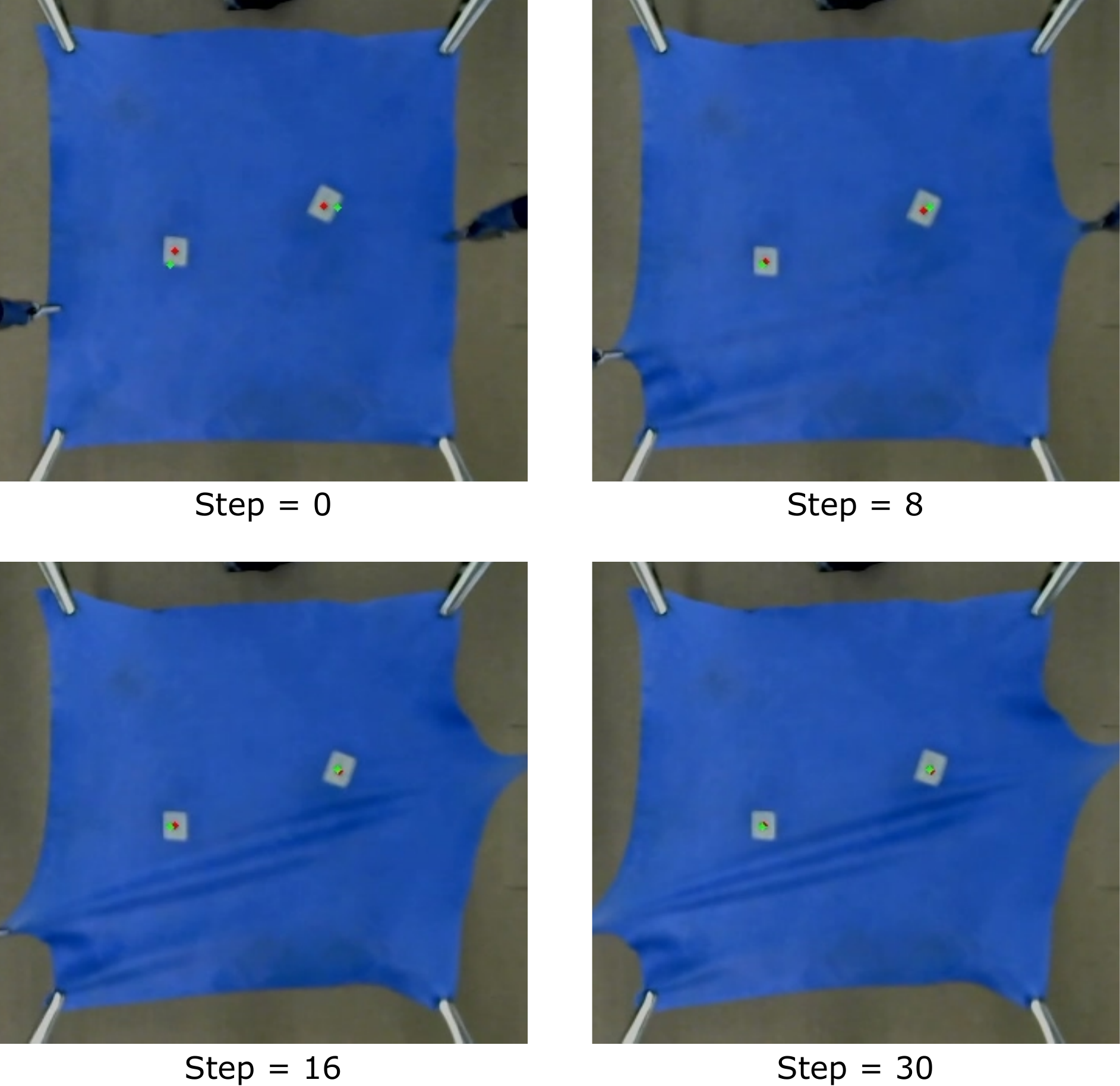}
\caption{A sequence of snapshots during a successful manipulation. The controlled points are tracked and colored in red, and the desired locations are marked as green.} \label{fig:snapshots}
\end{figure}

\begin{figure}[t]
\centering
\includegraphics[width=2.8in]{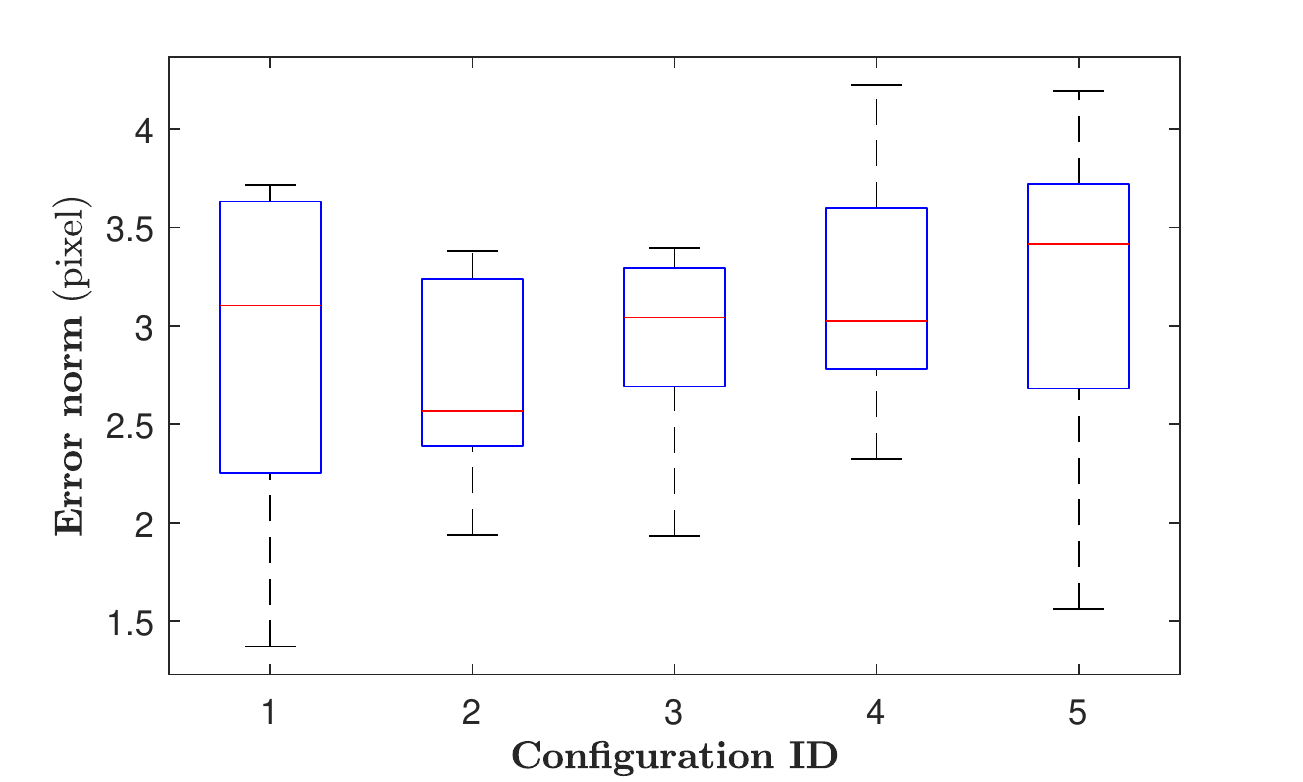}
\vspace{-0.1in}
\caption{Positioning error for 5 different configurations.} 
\vspace{-0.05in}
\label{fig:final_error}
\end{figure}

\begin{figure}
\centering
    \subfloat[\label{fig:error_t}]{%
      \includegraphics[width=3in]{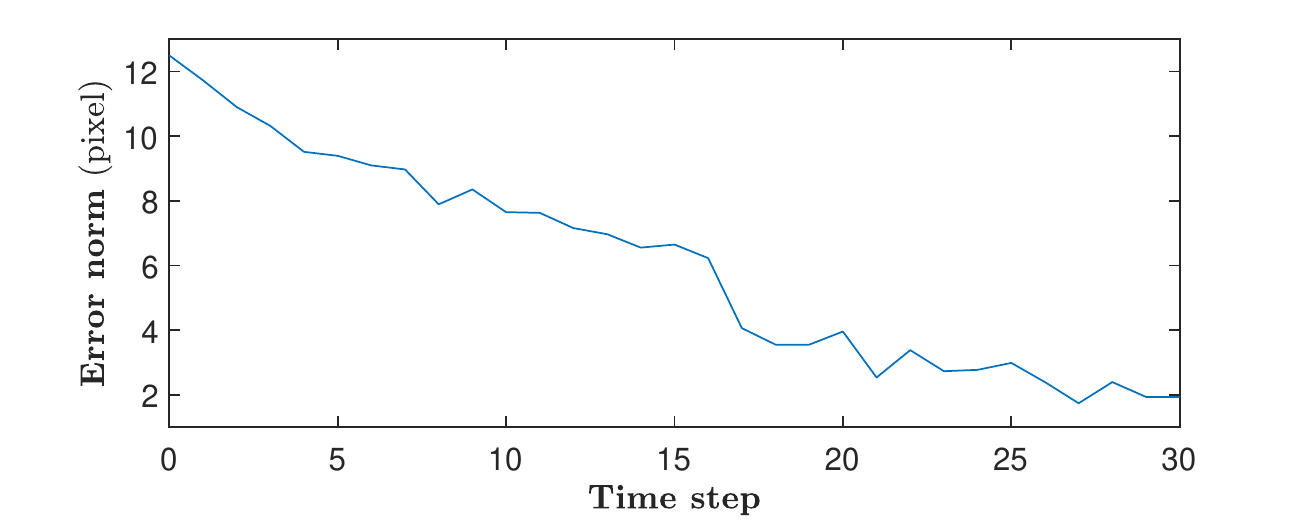}
    }\par
    \vspace{-0.15in}
    \subfloat[\label{fig:grasping_left}]{%
      \includegraphics[width=1.3in]{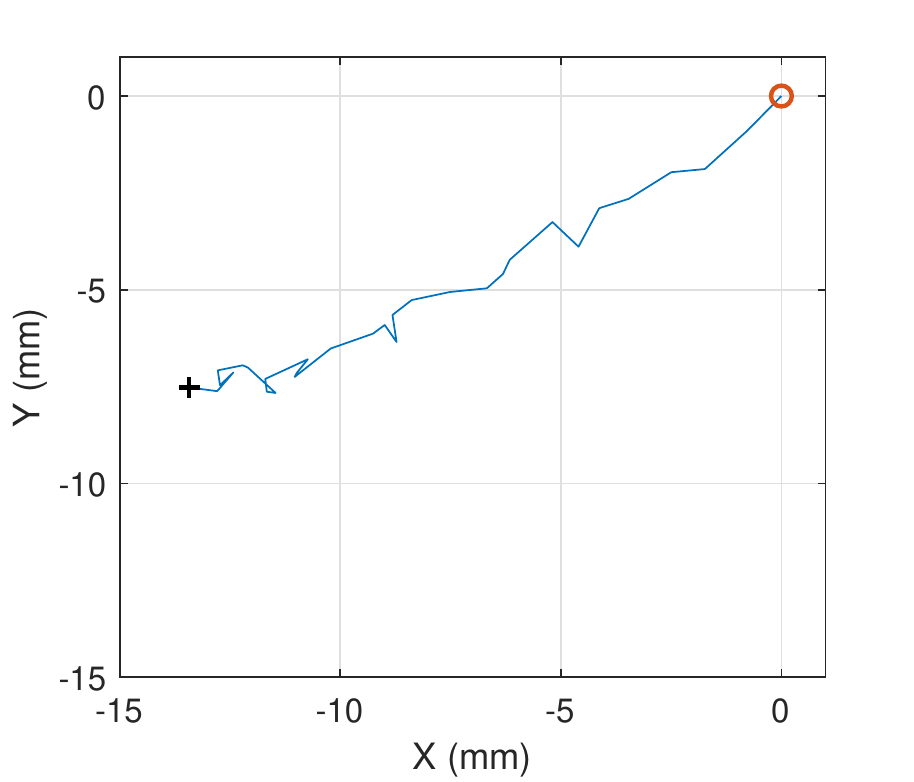}%
    }\hfil \hspace{-0.4in}
    \subfloat[\label{fig:grasping_right}]{%
      \includegraphics[width=1.3in]{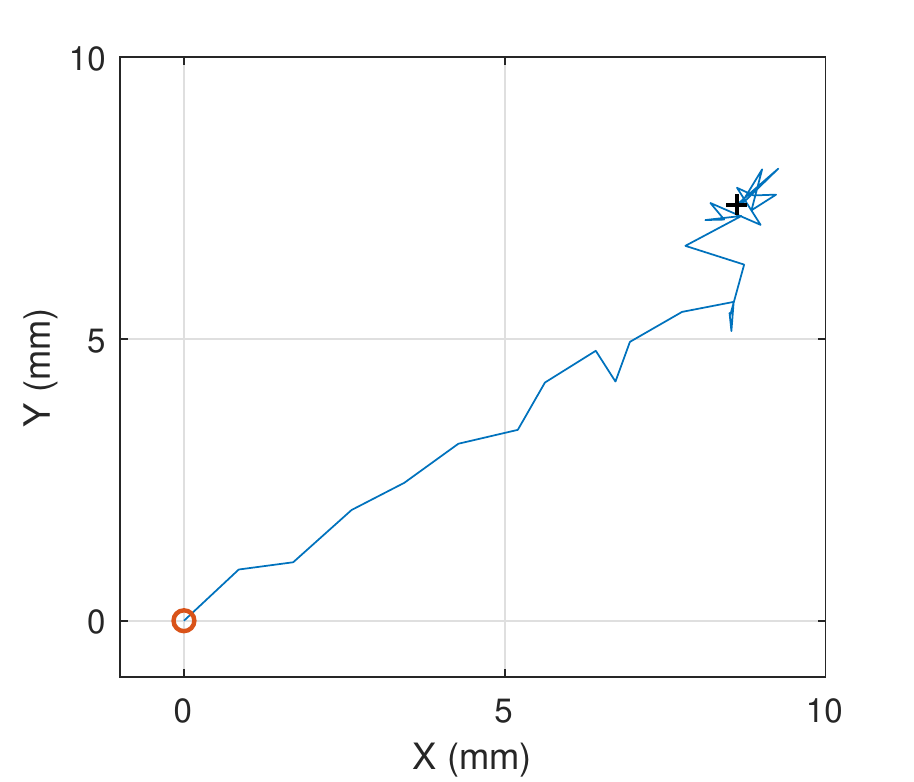}%
    }
    \caption{(a) Positioning error during a successful manipulation; (b) and (c) show the movement of the two grasping points in the 2D plane, respectively. Initial positions are aligned with the origin and marked by ``$\circ$", and the final positions after manipulation are marked by ``+".}
    \label{fig:error_time}
\end{figure}

\subsection{Results}
Fig.~\ref{fig:learning_curve} shows the learning curve for the SAC agents with and without state augmentation. The wall-clock time spent on each training instance is around 4.1 hours. The agent with state augmentation is able to reach a high average return (around 600) quickly after the training starts and ends up at around 900 after 20,000 steps, whereas the agent without state augmentation starts at a much lower value in the first evaluation, and its overall training process is very unstable and the average return reaches only around 550 at the end. {As a comparison, the hand-crafted policy achieves an average return of around 820 and a standard deviation of 441 (out of 10 episodes).} These results show that the proposed method is capable of learning a good policy in the FEM-based dynamic simulation environment without the need of increasing the iteration number. This potentially saves the time spent on simulation steps when solving the FEM system since a much smaller maximum number of iterations can be chosen (50 in our case), thus reducing the total training time.

Fig.~\ref{fig:grasping_displacement} shows the final displacement of the two grasping points in millimeters at the end of each trial on the real setup, which represents how large the deformation of the tissue is after the controlled points are aligned with the desired locations. The average planning time for the 5 configurations is 377 seconds.
Compared with other random selections, choosing the grasping points planned by the proposed method yields a relatively small displacement among the 10 trials. Although they are not always the smallest among all the trials, possibly due to the limitation of Bayesian optimization and the registration discrepancy between the real environment and the simulated one, it is reasonable to conclude that the planned grasping points are close to the optimal locations and that planning in the simulation environment leads to results transferable to the real world.

Fig.~\ref{fig:snapshots} shows a sequence of snapshots taken at certain time steps during a successful manipulation. As shown, the trained policy is able to place the controlled tissue points at the desired locations within 30 control steps. Fig.~\ref{fig:final_error} shows the localization error of the controlled points for each of the 5 different configurations. The trained policy is able to place the two controlled points at the desired locations with an error of around 3 pixels (approximately $1.3 \, \text{mm}$), and less than 4.5 pixels (approximately $2.0 \, \text{mm}$). Fig.~\ref{fig:error_time} shows the error in pixels squared against the time steps during one of the trials. The error converges to a steady state after around 25 control steps. Furthermore, we show through a two-way analysis of variance (ANOVA) that there is no significant difference of the positioning error both between each configuration groups and between using optimal or random grasping points, as shown in Table~\ref{tab:anova}. This indicates that the learned policy performs consistently well across different scenarios, even when using randomly chosen grasping points.

\begin{table}[]
\caption{ANOVA Results of Positioning Error with Regard to Grasping Points (Optimal/Random) and Configurations\label{tab:anova}}
\centering
\begin{tabular}{lrrrrr}
\hline
 \multicolumn{1}{l}{Source} & \multicolumn{1}{c}{Sum Sq.} & \multicolumn{1}{c}{d.f.} & \multicolumn{1}{c}{Mean Sq.} & \multicolumn{1}{c}{F} & \multicolumn{1}{c}{P-value} \\ \hline
 Grasping point & 0.2592 & 1 & 0.25922 & 0.57 & 0.4526 \\
 Configurations & 1.6431 & 4 & 0.41077 & 0.91 & 0.4666 \\
 Error & 19.8638 & 44 & 0.45145 & - & - \\
 Total & 21.7661 & 49 & - & - & - \\ \hline
\end{tabular}
\end{table}

\section{Discussion and Conclusion}
In this work, a sim-to-real learning and planning scheme is introduced for internal tissue point manipulation in surgeries. We demonstrate that a DRL agent with state augmentation is able to learn good manipulation policies in an FEM-based simulator without the need of increasing simulation time, and that preoperative planning in simulation can be achieved by utilizing the learned DRL policy. To ensure that the policy learned in the simulation is transferable to the real world, the domain randomization technique is utilized.
We report an average training time of 4.1 hours and a planning time of around 377 seconds on a CPU device. As a comparison, the average training time is 6.8 hours when the maximum number of iterations is doubled, in which case the agent using non-augmented MDP is also able to learn a reasonably good policy. However, this increases the training time by over 60\%, suggesting that the proposed method can save a significant amount of training time.

While the experimental setup in this work is simple, the proposed semi-autonomous learning and planning scheme is generally applicable to more complex real surgical scenarios where accurate positioning of tissue points is required, such as kidney cryoablation and breast brachytherapy, provided that a simulation environment exists. In addition, the proposed method is generally applicable to non-homogeneous tissue, as long as it can be modeled using FEM.

One limitation of this work is it assumes that the tissue is flat and only slightly warps during manipulation, while in practice the tissue surface is often irregular and highly wrapped. However, this can be addressed by ensuring an accurate 3D reconstruction of the tissue to build a realistic simulation environment when applying the proposed method to real surgeries. In addition, since in this work the controlled points are assumed to be always visually detectable for simplicity, which is usually not the case in real surgeries, our future research will incorporate the method described in \cite{afshar2022accurate} for internal tissue point localization. Furthermore, it is also possible that small local deformations at the grasping points may still cause damage to the tissue due to the application of large forces. Therefore, further work should be done on finding the optimal grasping points that minimize the force applied to the tissue. In our future work, we plan to extend the proposed method to breast manipulation for placing internal points at desired locations during breast brachytherapy, where the location of the internal points can be estimated based on the surface of the breast. {By utilizing the stereo camera, autonomous grasping without human intervention can be further achieved.}

Although the training and planning time are still high considering the scale of the problem, limiting its practical usage in real surgeries where the environment is more complicated, these can be further improved by utilizing GPUs for both FEM simulation and RL training. In addition, introducing parallelism by running multiple simulations at the same time and using parallel training and optimization algorithms can potentially reduce both the training and the planning time. Imitation learning techniques, such as pretraining the RL agent with behavior cloning from expert demonstrations can be incorporated to further accelerate training.


\bibliographystyle{IEEEtran}
\bibliography{references.bib}

\begin{thebibliography}{10}
\providecommand{\url}[1]{#1}
\csname url@samestyle\endcsname
\providecommand{\newblock}{\relax}
\providecommand{\bibinfo}[2]{#2}
\providecommand{\BIBentrySTDinterwordspacing}{\spaceskip=0pt\relax}
\providecommand{\BIBentryALTinterwordstretchfactor}{4}
\providecommand{\BIBentryALTinterwordspacing}{\spaceskip=\fontdimen2\font plus
\BIBentryALTinterwordstretchfactor\fontdimen3\font minus
  \fontdimen4\font\relax}
\providecommand{\BIBforeignlanguage}[2]{{%
\expandafter\ifx\csname l@#1\endcsname\relax
\typeout{** WARNING: IEEEtran.bst: No hyphenation pattern has been}%
\typeout{** loaded for the language `#1'. Using the pattern for}%
\typeout{** the default language instead.}%
\else
\language=\csname l@#1\endcsname
\fi
#2}}
\providecommand{\BIBdecl}{\relax}
\BIBdecl

\bibitem{chen2016virtual}
Z.~Chen, A.~Malpani, P.~Chalasani, A.~Deguet, S.~S. Vedula, P.~Kazanzides, and
  R.~H. Taylor, ``Virtual fixture assistance for needle passing and knot
  tying,'' in \emph{2016 IEEE/RSJ International Conference on Intelligent
  Robots and Systems (IROS)}.\hskip 1em plus 0.5em minus 0.4em\relax IEEE,
  2016, pp. 2343--2350.

\bibitem{chiu2021bimanual}
Z.-Y. Chiu, F.~Richter, E.~K. Funk, R.~K. Orosco, and M.~C. Yip, ``Bimanual
  regrasping for suture needles using reinforcement learning for rapid motion
  planning,'' in \emph{2021 IEEE International Conference on Robotics and
  Automation (ICRA)}.\hskip 1em plus 0.5em minus 0.4em\relax IEEE, 2021, pp.
  7737--7743.

\bibitem{richter2021autonomous}
F.~Richter, S.~Shen, F.~Liu, J.~Huang, E.~K. Funk, R.~K. Orosco, and M.~C. Yip,
  ``Autonomous robotic suction to clear the surgical field for hemostasis using
  image-based blood flow detection,'' \emph{IEEE Robotics and Automation
  Letters}, vol.~6, no.~2, pp. 1383--1390, 2021.

\bibitem{li2020super}
Y.~Li, F.~Richter, J.~Lu, E.~K. Funk, R.~K. Orosco, J.~Zhu, and M.~C. Yip,
  ``Super: A surgical perception framework for endoscopic tissue manipulation
  with surgical robotics,'' \emph{IEEE Robotics and Automation Letters},
  vol.~5, no.~2, pp. 2294--2301, 2020.

\bibitem{lu2021super}
J.~Lu, A.~Jayakumari, F.~Richter, Y.~Li, and M.~C. Yip, ``Super deep: A
  surgical perception framework for robotic tissue manipulation using deep
  learning for feature extraction,'' in \emph{2021 IEEE International
  Conference on Robotics and Automation (ICRA)}.\hskip 1em plus 0.5em minus
  0.4em\relax IEEE, 2021, pp. 4783--4789.

\bibitem{lin2022semantic}
S.~Lin, A.~J. Miao, J.~Lu, S.~Yu, Z.-Y. Chiu, F.~Richter, and M.~C. Yip,
  ``Semantic-super: A semantic-aware surgical perception framework for
  endoscopic tissue classification, reconstruction, and tracking,'' \emph{arXiv
  preprint arXiv:2210.16674}, 2022.

\bibitem{zhong2019dual}
F.~Zhong, Y.~Wang, Z.~Wang, and Y.-H. Liu, ``Dual-arm robotic needle insertion
  with active tissue deformation for autonomous suturing,'' \emph{IEEE Robotics
  and Automation Letters}, vol.~4, no.~3, pp. 2669--2676, 2019.

\bibitem{afshar2022model}
M.~Afshar, J.~Carriere, T.~Meyer, R.~Sloboda, S.~Husain, N.~Usmani, and
  M.~Tavakoli, ``A model-based multi-point tissue manipulation for enhancing
  breast brachytherapy,'' \emph{IEEE Transactions on Medical Robotics and
  Bionics}, 2022.

\bibitem{alambeigi2018toward}
F.~Alambeigi, Z.~Wang, Y.-h. Liu, R.~H. Taylor, and M.~Armand, ``Toward
  semi-autonomous cryoablation of kidney tumors via model-independent
  deformable tissue manipulation technique,'' \emph{Annals of biomedical
  engineering}, vol.~46, no.~10, pp. 1650--1662, 2018.

\bibitem{wu2019learning}
Y.~Wu, W.~Yan, T.~Kurutach, L.~Pinto, and P.~Abbeel, ``Learning to manipulate
  deformable objects without demonstrations,'' \emph{arXiv preprint
  arXiv:1910.13439}, 2019.

\bibitem{shin2019autonomous}
C.~Shin, P.~W. Ferguson, S.~A. Pedram, J.~Ma, E.~P. Dutson, and J.~Rosen,
  ``Autonomous tissue manipulation via surgical robot using learning based
  model predictive control,'' in \emph{2019 International Conference on
  Robotics and Automation}.\hskip 1em plus 0.5em minus 0.4em\relax IEEE, 2019,
  pp. 3875--3881.

\bibitem{tobin2017domain}
J.~Tobin, R.~Fong, A.~Ray, J.~Schneider, W.~Zaremba, and P.~Abbeel, ``Domain
  randomization for transferring deep neural networks from simulation to the
  real world,'' in \emph{2017 IEEE/RSJ international conference on intelligent
  robots and systems (IROS)}.\hskip 1em plus 0.5em minus 0.4em\relax IEEE,
  2017, pp. 23--30.

\bibitem{thananjeyan2017multilateral}
B.~Thananjeyan, A.~Garg, S.~Krishnan, C.~Chen, L.~Miller, and K.~Goldberg,
  ``Multilateral surgical pattern cutting in 2d orthotropic gauze with deep
  reinforcement learning policies for tensioning,'' in \emph{2017 IEEE
  International Conference on Robotics and Automation (ICRA)}.\hskip 1em plus
  0.5em minus 0.4em\relax IEEE, 2017, pp. 2371--2378.

\bibitem{tagliabue2020soft}
E.~Tagliabue, A.~Pore, D.~Dall’Alba, E.~Magnabosco, M.~Piccinelli, and
  P.~Fiorini, ``Soft tissue simulation environment to learn manipulation tasks
  in autonomous robotic surgery,'' in \emph{2020 IEEE/RSJ International
  Conference on Intelligent Robots and Systems (IROS)}.\hskip 1em plus 0.5em
  minus 0.4em\relax IEEE, 2020, pp. 3261--3266.

\bibitem{liu2021real}
F.~Liu, Z.~Li, Y.~Han, J.~Lu, F.~Richter, and M.~C. Yip, ``Real-to-sim
  registration of deformable soft tissue with position-based dynamics for
  surgical robot autonomy,'' in \emph{2021 IEEE International Conference on
  Robotics and Automation (ICRA)}.\hskip 1em plus 0.5em minus 0.4em\relax IEEE,
  2021, pp. 12\,328--12\,334.

\bibitem{wada1998indirect}
T.~Wada, S.~Hirai, and S.~Kawamura, ``Indirect simultaneous positioning
  operations of extensionally deformable objects,'' in \emph{Proceedings. 1998
  IEEE/RSJ International Conference on Intelligent Robots and Systems.
  Innovations in Theory, Practice and Applications (Cat. No. 98CH36190)},
  vol.~2.\hskip 1em plus 0.5em minus 0.4em\relax IEEE, 1998, pp. 1333--1338.

\bibitem{faure2012sofa}
F.~Faure, C.~Duriez, H.~Delingette, J.~Allard, B.~Gilles, S.~Marchesseau,
  H.~Talbot, H.~Courtecuisse, G.~Bousquet, I.~Peterlik \emph{et~al.}, ``Sofa: A
  multi-model framework for interactive physical simulation,'' in \emph{Soft
  tissue biomechanical modeling for computer assisted surgery}.\hskip 1em plus
  0.5em minus 0.4em\relax Springer, 2012, pp. 283--321.

\bibitem{tagliabue2020biomechanical}
E.~Tagliabue, D.~Dall’Alba, E.~Magnabosco, I.~Peterlik, and P.~Fiorini,
  ``Biomechanical modelling of probe to tissue interaction during ultrasound
  scanning,'' \emph{International Journal of Computer Assisted Radiology and
  Surgery}, vol.~15, no.~8, pp. 1379--1387, 2020.

\bibitem{haarnoja2018soft}
T.~Haarnoja, A.~Zhou, K.~Hartikainen, G.~Tucker, S.~Ha, J.~Tan, V.~Kumar,
  H.~Zhu, A.~Gupta, P.~Abbeel \emph{et~al.}, ``Soft actor-critic algorithms and
  applications,'' \emph{arXiv preprint arXiv:1812.05905}, 2018.

\bibitem{altman1992closed}
E.~Altman and P.~Nain, ``Closed-loop control with delayed information,''
  \emph{ACM sigmetrics performance evaluation review}, vol.~20, no.~1, pp.
  193--204, 1992.

\bibitem{csaji2008value}
B.~C. Cs{\'a}ji and L.~Monostori, ``Value function based reinforcement learning
  in changing markovian environments.'' \emph{Journal of Machine Learning
  Research}, vol.~9, no.~8, 2008.

\bibitem{stable-baselines3}
\BIBentryALTinterwordspacing
A.~Raffin, A.~Hill, A.~Gleave, A.~Kanervisto, M.~Ernestus, and N.~Dormann,
  ``Stable-baselines3: Reliable reinforcement learning implementations,''
  \emph{Journal of Machine Learning Research}, vol.~22, no. 268, pp. 1--8,
  2021. [Online]. Available: \url{http://jmlr.org/papers/v22/20-1364.html}
\BIBentrySTDinterwordspacing

\bibitem{kazanzides-chen-etal-icra-2014}
P.~Kazanzides, Z.~Chen, A.~Deguet, G.~S. Fischer, R.~H. Taylor, and S.~P.
  DiMaio, ``An open-source research kit for the da vinci surgical system,'' in
  \emph{IEEE Intl. Conf. on Robotics and Auto. (ICRA)}, Hong Kong, China, 2014,
  pp. 6434--6439.

\bibitem{santos2021new}
A.~E.~C. Santos, F.~N. Linhares, M.~C. A.~M. Leite, V.~A. Esc{\'o}cio, and
  R.~C.~R. Nunes, ``A new device to simulate the performance of rubber dams for
  dental applications,'' \emph{Polymer Testing}, vol.~94, p. 107043, 2021.

\bibitem{singh2021mechanical}
G.~Singh and A.~Chanda, ``Mechanical properties of whole-body soft human
  tissues: a review,'' \emph{Biomedical Materials}, 2021.

\bibitem{afshar2022accurate}
M.~Afshar, J.~Carriere, H.~Rouhani, T.~Meyer, R.~S. Sloboda, S.~Husain,
  N.~Usmani, and M.~Tavakoli, ``Accurate tissue deformation modeling using a
  kalman filter and admm-based projective dynamics,'' \emph{IEEE/ASME
  Transactions on Mechatronics}, 2022.

\end{thebibliography}

\end{document}